\documentclass[conference]{IEEEtran}
\IEEEoverridecommandlockouts

\usepackage{cite}
\usepackage{amsmath,amssymb,amsfonts}
\usepackage{algorithm}
\usepackage{algpseudocode}
\usepackage{graphicx}
\usepackage[font=small,labelfont=bf]{caption}
\usepackage{textcomp}
\usepackage{enumitem}
\usepackage[square,numbers]{natbib}
\usepackage{xcolor}
\usepackage{makecell}
\usepackage{multirow}
\def\BibTeX{{\rm B\kern-.05em{\sc i\kern-.025em b}\kern-.08em
    T\kern-.1667em\lower.7ex\hbox{E}\kern-.125emX}}
\algnewcommand{\LineComment}[1]{\State \(\triangleright\) #1}

\usepackage{fancyhdr,lipsum}
\fancypagestyle{firstpage}{
  \fancyhf{}
  \fancyhead[C]{To appear at the 2020 International Joint Conference on Neural Networks (IJCNN), Glasgow, Scotland, July 2020.}
  \fancyfoot[C]{\thepage}
}

\begin{document}

\title{FasTrCaps: An Integrated Framework for Fast yet Accurate Training of Capsule Networks\\
}


\author{\IEEEauthorblockN{Alberto Marchisio$^{1,*}$\thanks{*These authors contributed equally to this work.}, Beatrice Bussolino$^{2,*}$, Alessio Colucci$^{1,*}$, Muhammad Abdullah Hanif $^1$,\\Maurizio Martina$^2$, Guido Masera$^2$,Muhammad Shafique$^1$}
\IEEEauthorblockA{\textit{$^1$Technische Universit{\"a}t Wien, Vienna, Austria}} 
\IEEEauthorblockA{\textit{$^2$Politecnico di Torino, Turin, Italy}\\
Email: \{alberto.marchisio,alessio.colucci,muhammad.hanif,muhammad.shafique\}@tuwien.ac.at\\
\{beatrice.bussolino,maurizio.martina,guido.masera\}@polito.it\\
}\vspace*{-20pt}}

\maketitle
\thispagestyle{firstpage}

\begin{abstractsize}
\begin{abstract}
Recently, Capsule Networks (CapsNets) have shown improved performance compared to the traditional Convolutional Neural Networks (CNNs), by encoding and preserving spatial relationships between the detected features in a better way. This is achieved through the so-called Capsules (i.e., groups of neurons) that encode both the instantiation probability and the spatial information. However, one of the major hurdles in the wide adoption of CapsNets is their gigantic training time, which is primarily due to the relatively higher complexity of their new constituting elements that are different from CNNs.

In this paper, we implement different optimizations in the training loop of the CapsNets, and investigate how these optimizations affect their training speed and the accuracy. Towards this, we propose a novel framework \textit{FasTrCaps} that integrates multiple lightweight optimizations and a novel learning rate policy called \textit{WarmAdaBatch} (that jointly performs \textit{warm restarts} and \textit{adaptive batch size}), and steers them in an appropriate way to provide high training-loop speedup at minimal accuracy loss. We also propose  \textit{weight sharing} for capsule layers. The goal is to reduce the hardware requirements of CapsNets by removing unused/redundant connections and capsules, while keeping high accuracy through tests of different learning rate policies and batch sizes. We demonstrate that one of the solutions generated by the \textit{FasTrCaps} framework can achieve 58.6\% reduction in the training time, while preserving the accuracy (even 0.12\% accuracy improvement for the MNIST dataset), compared to the CapsNet by Google Brain~\cite{sabourdynamic2017}. Moreover, the Pareto-optimal solutions generated by \textit{FasTrCaps} can be leveraged to realize trade-offs between training time and achieved accuracy. We have open-sourced our framework on GitHub\footnote{\url{https://github.com/Alexei95/FasTrCaps}}.
\end{abstract}

\begin{IEEEkeywords}
Machine Learning, Capsule Networks, Training, Accuracy, Efficiency, Performance, Weight Sharing, Decoder, Batch Sizing, Adaptivity.
\end{IEEEkeywords}
\end{abstractsize}


\section{Introduction}
The development of Deep Neural Networks (DNNs), especially the Convolutional Neural Networks (CNNs), has experienced a dramatic increase in the past decade, leading to many different architectures~\cite{Marchisio2019DL4EC}\cite{Shafique2018NextGenML}. A key problem is the optimization of CNNs and their hyper-parameters, for which, most of the fine-tuning optimizations (e.g., choosing the training policy, the learning rate, the optimizer, etc.) are repetitive and time-consuming, because every change must be tested with many epochs of training and repeated many times to have certain statistical significance. This is significantly worsened when increasing the CNN complexity, which leads to a more demanding compute effort to find the right set of optimizations.

Different methods have been explored in the literature to reduce the training time of CNNs, such as \textit{one-cycle policy}~\cite{Smith2018}, \textit{warm restarts}~\cite{Loshchilov}, and \textit{adaptive batch sizing}~\cite{De}\cite{Devarakonda}\cite{Goyal}. However, CNNs have a key limitation: they do not retain information on the spatial correlation between the detected features. This effect causes poor network performances in terms of accuracy when the object to be recognized is rotated, has a different orientation, or presents any other geometrical variation. Currently, this problem is solved by training CNNs on expanded large-sized datasets, that include also transformed and modified objects. However, wider datasets lead to much longer training times, which not only pose delays in the DNN development cycle, but also require (1) super-costly training machines like Nvidia's DGX-2 or DGX Superpod rendering it unaffordable for many developers/organizations, or (2) outsourcing to third party cloud services that can compromise privacy and security requirements~\cite{Shafique2020RobustML}\cite{Zhang2019RobustML}. Hence, both advanced DNN architectures and fast training techniques are necessary to mitigate the above challenges.

Capsule Networks (CapsNets) by Google~\cite{sabourdynamic2017} aim at overcoming the limitations of CNNs w.r.t. preserving the spatial correlation between the detection features through the following two means: (1) by substituting single neurons with the so-called \textit{capsules} (i.e., groups of neurons); and (2) using the \textit{dynamic routing} algorithm between the capsule layers that provides the ability to encode both the instantiation probability of an object and its instantiation parameters (width, orientation, skew, etc.). \textit{However, one of the major hurdles in the wide-scale adoption of CapsNets is their gigantic training time, which is primarily due to the higher complexity of their new constituting elements.}

\textbf{Motivational Case Study and Target Research Problem:}

\begin{figure}[h]
	\centering
	\includegraphics[width=\linewidth]{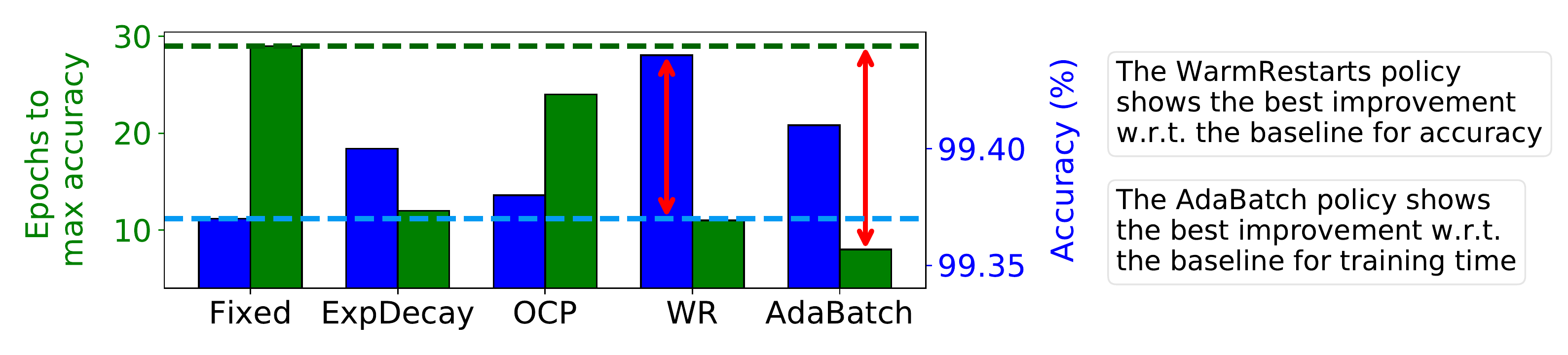}
	\caption{Optimization potentials when considering CapsNet.}
	\label{fig:optimization_potential}
\end{figure}

Our analyses in Fig.~\ref{fig:optimization_potential} show the available improvement potential and opportunities for the CapsNets, when different optimizations like \textit{Exponential Decay} (ExpDecay), \textit{One-Cycle-Policy} (OCP)~\cite{Smith2018}, \textit{Warm Restarts} (WR)~\cite{Loshchilov} and \textit{Adaptive Batch Size} (AdaBatch)~\cite{De} are applied. These optimizations result in a higher accuracy and a lower number of epochs to reach the maximum accuracy, w.r.t. the baseline (Fixed) learning rate policy. This motivates us to tailor these optimizations towards the new features of CapsNets, and implement them in an integrated framework to train CapsNets in a fast yet accurate manner, while also reducing the number of CapsNet parameters, for instance, through \textit{weight sharing}.

\textbf{Our Novel Contributions and Concept Overview [Fig. 2]:}
We present \textit{FasTrCaps}, a framework which employs different optimization methods for significantly reducing the training time and the number of parameters of CapsNets, while preserving or improving their accuracy\footnote{A previous version of this work is available in~\cite{Marchisio2019XTrainCaps}.}. 

\begin{figure}[h]
	\centering
	\includegraphics[width=\linewidth]{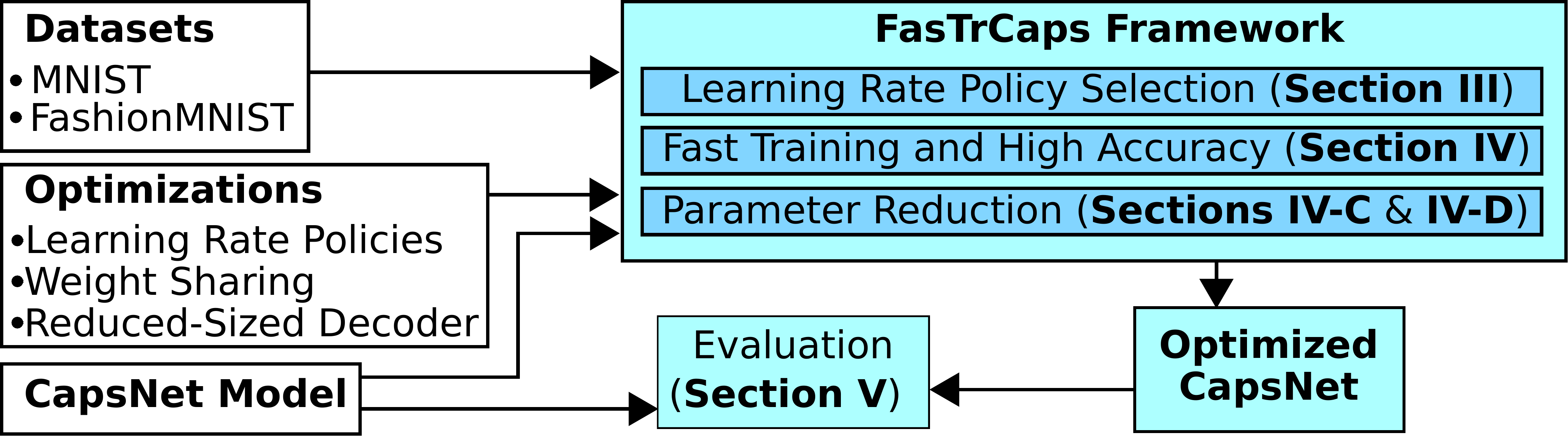}
	\caption{An overview of our novel contributions in this paper.}
	\label{fig:novel_contributions}
\end{figure}

\begin{figure*}[t]
\vspace*{0mm}
    \centering
    \includegraphics[width=\textwidth]{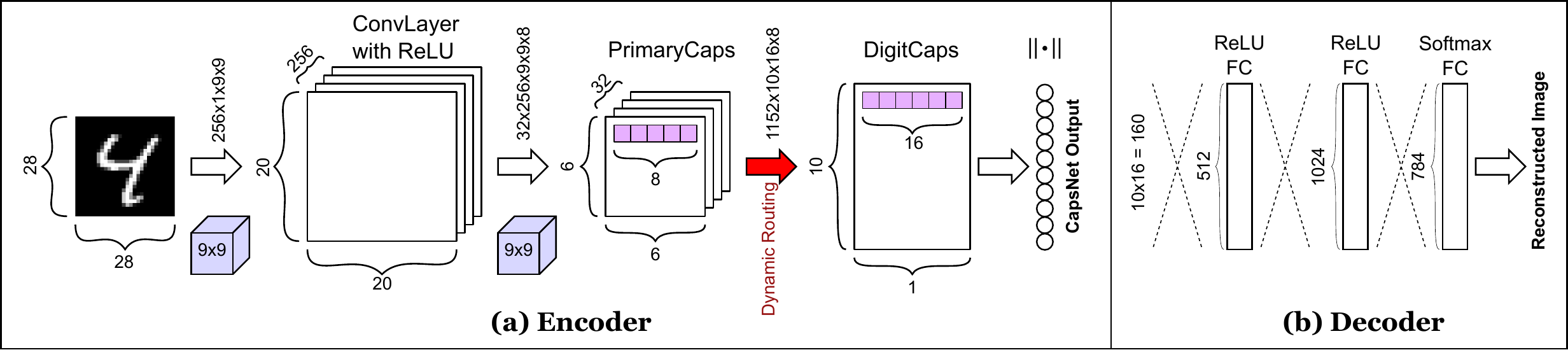}
    \caption{Architectural view of a CapsNet illustrating both the Encoder and Decoder parts.}
    \label{fig:capsnet}
\end{figure*}

\textbf{The key contributions are:}
\begin{itemize}[leftmargin=*]
    \item Tailoring different learning rate policies (like \textit{one-cycle policy} or \textit{warm restarts}) to specialize them for the CapsNet structure, and analyzing their efficiency in the CapsNet training loop vs. the corresponding training time (\textbf{Section~\ref{sec:motivational_analysis}}).
    \item A novel training framework, \textit{FasTrCaps}, which accelerates the training of CapsNets by integrating the above-described optimizations (like \textit{warm restarts}, \textit{adaptive batch size} and \textit{weight sharing}) in an automated flow specialized to the structure and training flow of the CapsNets (i.e., considering the capsules and the coupling between capsule layers) (\textbf{Section~\ref{sec:methodology}}).
    \item Parameter reduction via \textit{weight sharing} and reducing the size of the decoder for CapsNets by removing its unused connections. These optimizations reduce the number of parameters by more than 15\% (\textbf{Section~\ref{sec:decoder} \&~\ref{sec:weight_sharing}}).
    \item Evaluation of the FasTrCaps framework on the MNIST and Fashion-MNIST datasets to stay compliant with the experimental setup of~\cite{sabourdynamic2017} (\textbf{Section~\ref{sec:evaluation}}).
    \item Open-source release of our FasTrCaps framework, for reproducibility and to facilitate further research and development in the area of accelerated training of CapsNets, at \url{https://github.com/Alexei95/FasTrCaps}.
\end{itemize}

Another key benefit of our \textit{FasTrCaps} framework is that it provides multiple Pareto-optimal solutions that can be leveraged to trade-off training time reductions without (or in some cases with a user-tolerable) accuracy loss. For instance, in our experimental evaluations, \textit{FasTrCaps} provides a solution that can reduce the training time of CapsNets by 58.6\% while providing a slight increase in its accuracy (i.e. 0.12\%). Another solution provides 15\% reduction in the number of parameters of the CapsNet without affecting its accuracy. Note, our methodology may also be beneficial for other complex CNNs, as it enables integration of available optimizations for training.

Before proceeding to the technical sections, we present an overview of CapsNets and the learning rate policies in \textbf{Section~\ref{sec:background}}, to a level of detail necessary to understand our contributions.

\section{Background and Related Work}
\label{sec:background}
\subsection{CapsNets}

The Capsule introduced by Hinton~\cite{hintontransforming2011} denotes a group of neurons encoding both the instantiation probability of an object and the spatial information. The works in~\cite{sabourdynamic2017} and~\cite{hintonmatrix2018} introduced two architectures based on capsules, tested on the MNIST dataset~\cite{mnist}, with a classification accuracy aligned to the state-of-the-art traditional CNNs for the same application. The CapsNet that we use in our work corresponds to the model of~\cite{sabourdynamic2017}, as shown in Fig.~\ref{fig:capsnet}. The layers constituting the encoder are:

\begin{itemize}[leftmargin=*]
    \item \textbf{ConvLayer:} an initial convolutional layer.
    \item \textbf{PrimaryCaps:} a layer that transforms the scalar numbers of ConvLayer in vectors.
    \item \textbf{DigitCaps:} a layer that performs dynamic routing and computes the output probabilities.
\end{itemize}

The encoding network is followed by a decoder, composed of three fully-connected layers, which output the reconstructed image. The loss computed on the reconstructed image (i.e., the \textit{Reconstruction Loss}) directs the capsules of the DigitCaps layer to encode the instantiation parameters of the object.

\begin{figure*}[h]
    \centering
    \vspace*{0mm}
    \includegraphics[width=\textwidth]{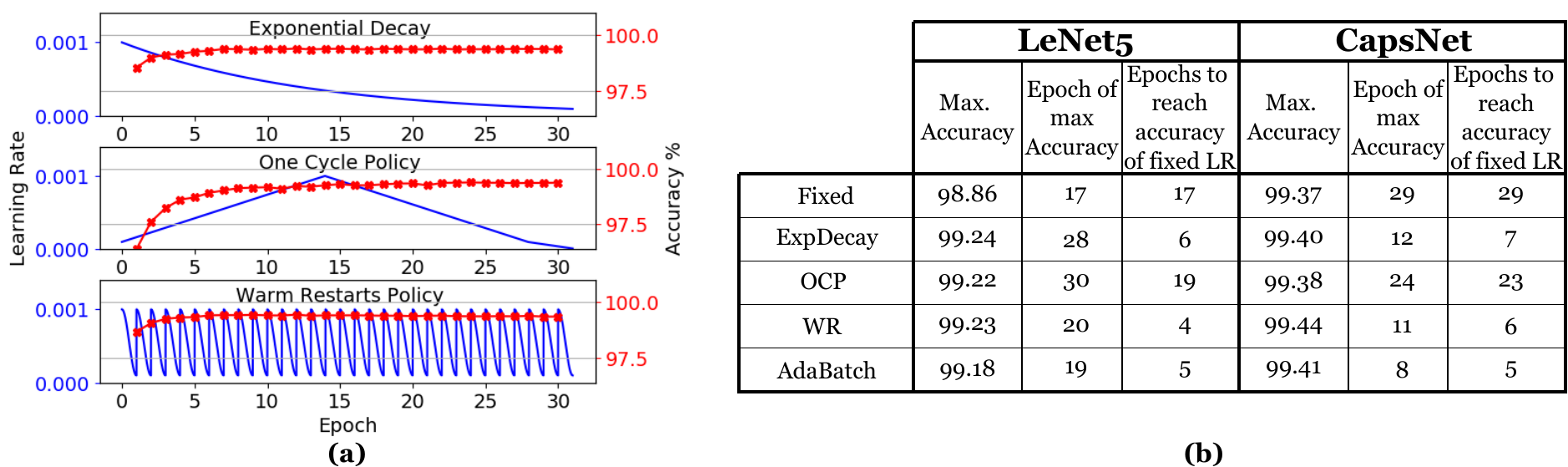}
    \caption[\textbf{(a)} In blue, how the learning rate changes when different policies like \textit{exponential decay}, \textit{one-cycle- policy} and \textit{warm restarts} are applied. In red, the accuracy reached by the CapsNet at every epoch of training with the corresponding learning rate policy applied. \textbf{(b)} A table summarizing the comparative differences between the LeNet5 and the CapsNet, when the same learning rate policies are applied, but considering appropriate functional enhancements (without violating their optimization function and flow) required to employ these policies for CapsNets. For each architecture, different columns of the table show: the maximum reached accuracy, training epochs to reach the maximum accuracy and the training epochs to reach the same accuracy of the network when fixed learning rate policy is used.]{\textbf{(a)} In blue, how learning rate changes when different learning rate policies like exponential decay, one cycle policy and \textit{warm restarts} are applied. In red, the accuracy reached by CapsNet at every epoch of training with the corresponding learning rate policy applied. \textbf{(b)} A table summarizing the comparative differences between LeNet5 and CapsNet when the same learning rate policies are applied. For each architecture, different columns of the table show, from left to right, the maximum reached accuracy, the training epochs\footnotemark[2] to reach the maximum accuracy, and the training epochs to reach the same accuracy of the network when the fixed learning rate policy is used.}
    \label{fig:policies}
\end{figure*}

Compared to traditional CNNs, CapsNets have shown high robustness against adversarial attacks~\cite{Michels2019VulnerabilityCaps} and affine transformations~\cite{Marchisio2019CapsAttacks}. These results contribute towards employing CapsNets in safety-critical applications. 
Moreover, from the computational perspective, CapsNets can be efficiently executed on IoT/Edge devices with the usage of specialized hardware~\cite{Marchisio2019CapStore}\cite{Marchisio2019CapsAcc}. Further energy savings can be achieved by applying quantization~\cite{Marchisio2020qcapsnets} or approximate multipliers~\cite{Marchisio2020ReDCaNe}.

\subsection{An Overview of Two Key Learning Rate Policies}
The learning rate (LR) is a relevant hyperparameter for the fast convergence during the training of a neural network. With a wide learning rate the optimization process may stop in a local minima or diverge, while a low learning rate can lead to a very slow convergence~\cite{bache}\cite{Bengio2012}\cite{breuel}. Given the difficulty of the choice of the best value for a constant learning rate, dynamic learning rate policies are often adopted, consisting in varying the learning rate during the training~\cite{wu}.

\textbf{One-Cycle Policy~\cite{Smith2017b}\cite{Smith2015}\cite{Smith2017}}: This method consists of three phases of training. In phase-1, the learning rate is linearly increased from a minimum to a maximum value in an optimal range. In phase-2, the learning rate is symmetrically decreased. In phase-3, the learning rate must be annealed to a very low value in a small fraction of the last steps. Equation~\ref{eq:onecycle} reports the formulas of the three phases of the \textit{one-cycle policy}, where $ts$ is the training step, $TS$ is the total number of steps in the training epochs, $lr_{min}$ and $lr_{max}$ are the learning rate range boundaries. Saddle points in the loss function slow down the training process, since the gradients in these regions have smaller values. Increasing the learning rate helps to faster traverse the saddle points.

\begin{equation} \label{eq:onecycle}
\resizebox{.91\linewidth}{!}{%
$
\begin{cases}
lr = lr_{min} + ts \cdot \frac{lr_{max}-lr_{min}}{0.45 \cdot TS} & 0<ts<0.45 \cdot TS $~~~~~~phase-1$ \\ 
lr = lr_{min} + (ts - 0.9 \cdot TS) \cdot \frac{lr_{min}-lr_{max}}{0.45 \cdot TS} & 0.45TS<ts<0.9TS $~~phase-2$ \\ 
lr = lr_{min} - 9 \cdot \frac{lr_{min}}{TS} \cdot (ts-0.9 \cdot TS) & 0.9 \cdot TS<ts<TS $~~~~~phase-3$
\end{cases}
$
}
\end{equation}

\textbf{Warm Restarts}: In the Stochastic Gradient Descent with Warm Restarts~\cite{Loshchilov} (aka \textit{warm restarts}), the learning rate is initialized to a maximum value and then it is decreased with cosine annealing until reaching the lower bound of a chosen interval. When the learning rate reaches the minimum value, it is again set to the maximum value, realizing a step function. The cosine annealing function is given in Equation~\ref{eq:warmrestarts}, where $lr_{min}$ and $lr_{max}$ are the learning rate range boundaries, $ts$ is the training step, $T_i$ is the number of training steps for each cycle. When $ts=Ti$, $ts$ is set to $0$ and the cycle starts again. This process is repeated cyclically during the whole training time, where the cycle period needs to be set properly to optimize the training time and accuracy. Increasing the learning rate step-wise emulates a warm restart of the network and encourages the model to step out from possible local minima or saddle points.

\begin{equation} \label{eq:warmrestarts}
    lr = lr_{min} + \frac{1}{2}\left( lr_{max} - lr_{min} \right) \left( 1 + cos \left( \pi \cdot \frac{ts}{T_i}\right)\right)
\end{equation}

Recently, a novel warm restart technique was proposed in~\cite{Mishra}, where the learning rate is decreased with a polynomial function after each restart.

\subsection{Adaptive Batch Size} 
Training a DNN with a small batch size can provide a faster convergence~\cite{JasKen17Three}\cite{Masters}, while a larger batch size allows to have an higher data parallelism and, consequently, high computational efficiency. Hence, many authors have studied methods to increase the batch size with fixed policies~\cite{Babanezhad}\cite{Daneshmand} or following an adaptive criterion, with the so-called \textit{Adaptive Batch Size}~\cite{De}\cite{Devarakonda}\cite{Goyal}. 

\begin{figure*}[t]
\vspace*{0mm}
    \centering
    \begin{minipage}[c]{.72\linewidth}
    \begin{figure}[H]
    \includegraphics[width=0.9\textwidth]{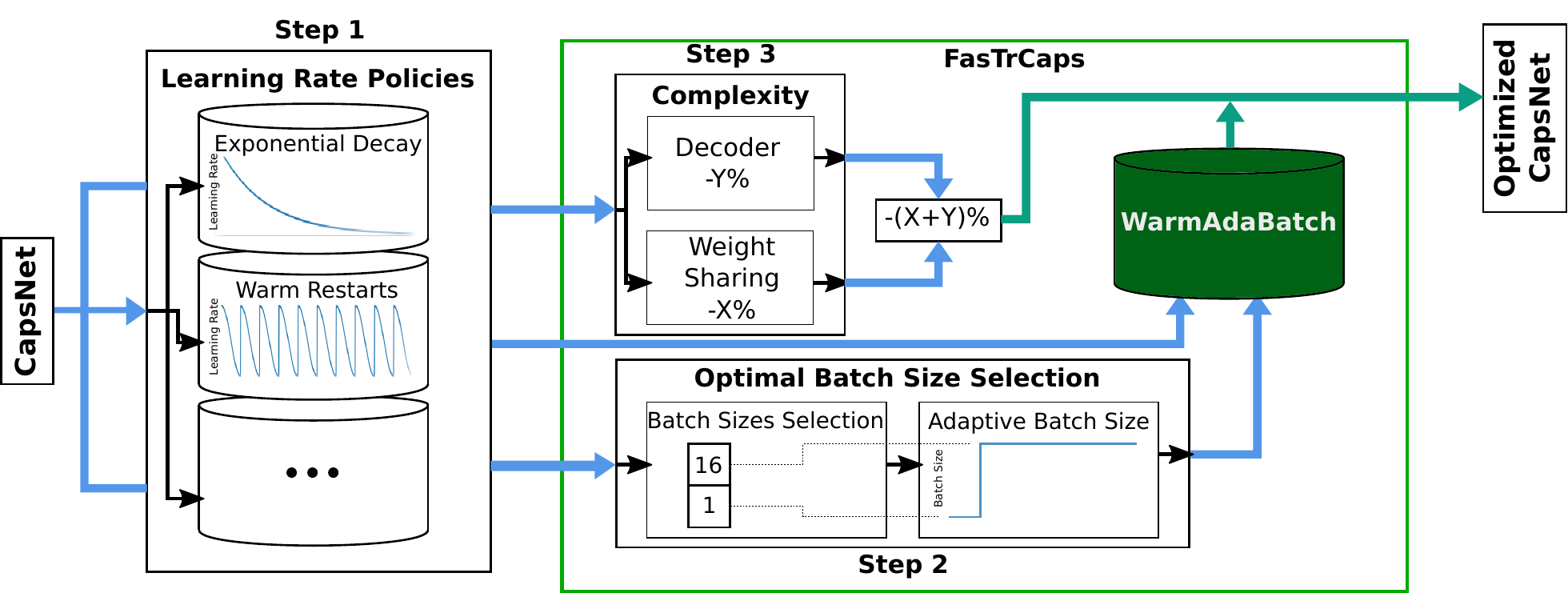}
    \caption{FasTrCaps processing flow: the CapsNet at the input goes through the different stages of optimization in parallel, to search for the right learning rate policy, batch size and complexity reduction, obtaining at the output the Optimized CapsNet, based on the optimization criteria chosen by the user.}
    \label{fig:workflow}
    \end{figure}
    \end{minipage}
    \hfill
    \begin{minipage}[c]{.25\linewidth}
    \begin{figure}[H]
    \vspace*{3mm}
    \includegraphics[width=\textwidth]{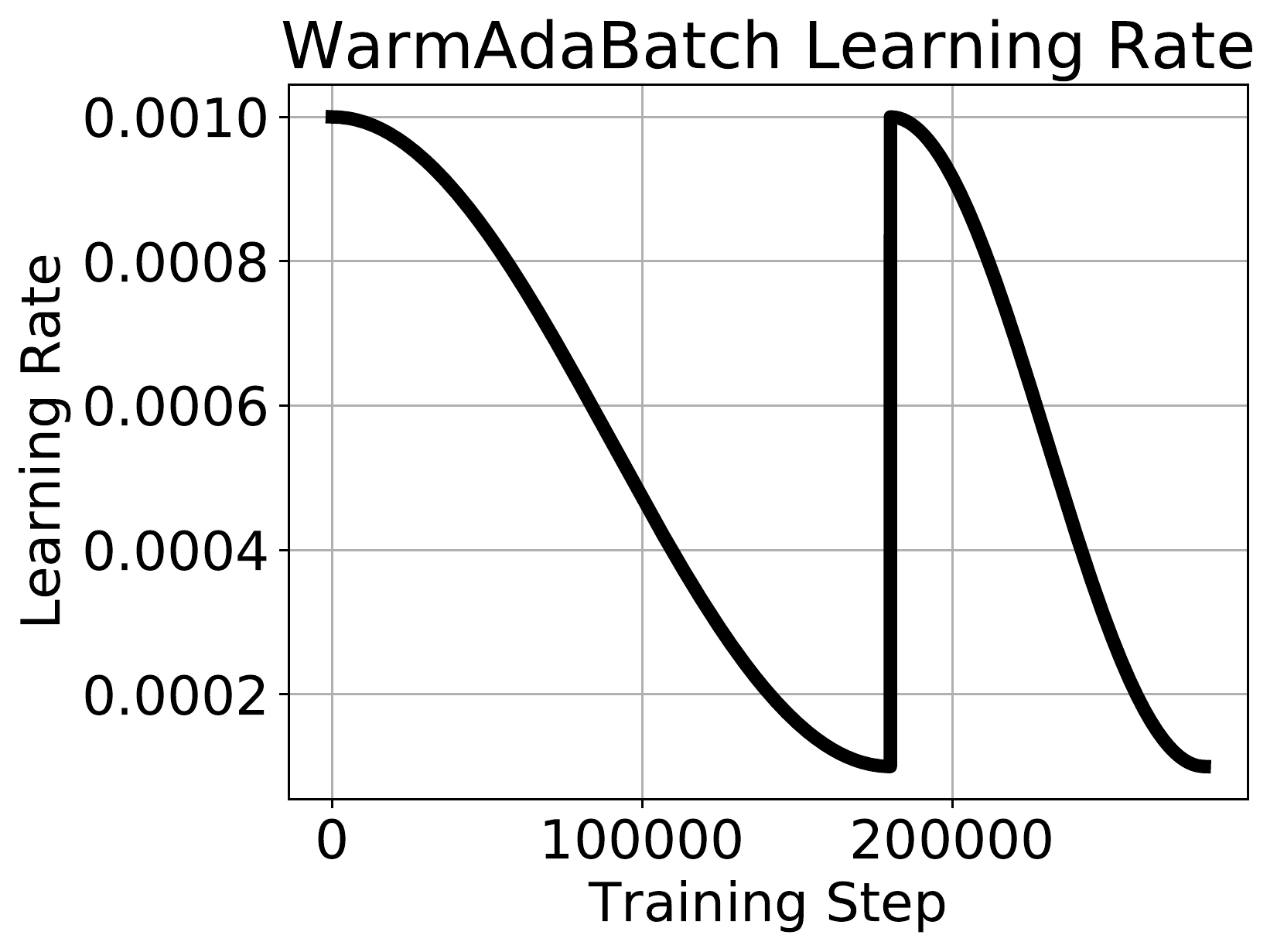}
    \vspace*{0.5mm}
    \caption{Learning rate policy for our \textit{WarmAdaBatch}.}
    \label{fig:learning_rate_warmadabatch}
    \end{figure}
    \end{minipage}
\end{figure*}
\footnotetext[2]{Training times for a single epoch are different between LeNet5 and CapsNet: the former takes an average of 17 seconds using the fixed learning rate, while the latter takes 49 seconds.}
\section{Analyzing the Efficiency of Learning Rate Policies for CapsNets}
\label{sec:motivational_analysis}
The techniques described in Section~\ref{sec:background} have been tailored for traditional CNNs, to improve their performances in terms of accuracy and training time. This section aims to customize different learning rate policies and batch size selection for training the CapsNets considering the multidimensional capsules and their cross-coupling, and to study whether and how much these policies are effective. 
\textit{Since the traditional neurons of the CNNs are replaced by capsules in the CapsNets, the number of parameters (weights and biases) to be trained is huge.}


For this purpose, we implemented different state-of-the-art learning rate policies for the training loop of the CapsNet, such that \textit{these techniques are enhanced for the capsule structures and relevant parameters of the CapsNet} (see discussion in Section~\ref{sec:methodology}). Fig.~\ref{fig:policies} (a) shows the learning rate changes for different techniques and how the accuracy of the CapsNet for the MNIST dataset varies accordingly. More detailed results of our analyses, including the comparisons with the LeNet5, are reported in Fig.~\ref{fig:policies} (b).

\noindent
From this analysis, we derive the following \textbf{key observations}:
\begin{enumerate}[leftmargin=*]
\item The \textit{warm restarts} technique is the most promising one because it allows to reach the same accuracy (99.37\%) as the CapsNet with a fixed learning rate, while providing a reduction of 79.31\% in the training time.
\item A more extensive training with \textit{warm restarts} leads to to an accuracy improvement of 0.07\%.
\item The \textit{adaptive batch size} shows similar improvements in terms of accuracy (99.41\%) and training epochs.
\item The first epochs with smaller batch sizes execute relatively longer when compared to the ones with bigger batch sizes.
\end{enumerate}


\section{FasTrCaps: Our Framework to Accelerate the Training of CapsNets}
\label{sec:methodology}


Training a CapsNet consists of a multi-objective optimization problem, because our scope is to maximize the accuracy, while minimizing the training time and the network complexity. A comprehensive processing flow of our \textit{FasTrCaps} framework is shown in Fig.~\ref{fig:workflow}. Before describing how to integrate different optimizations in an automated training methodology and how to generate the optimized CapsNet at the output (Section~\ref{sec:output_generation}), we present how these optimizations have been implemented with enhancements for the CapsNets, which is necessary to realize an integrated training framework.

\subsection{Learning Rate Policies for CapsNets}
\label{sec:learning_rate_capsnets}
The first parameter analyzed to improve the training process of CapsNets is the learning rate. The optimal learning rate range is evaluated within the range boundaries 0.0001 and 0.001. For our framework, we use the following parameters in these learning rate policies:
\begin{itemize}[leftmargin=*]
    \item \textbf{Fixed learning rate}: 0.001
    \item \textbf{Exponential decay}: starting value 0.001, decay rate 0.96, decay steps 2,000: $lr = lr_0 \cdot 0.96^{current\_step / 2,000}$
    \item \textbf{One cycle policy}: lower bound 0.0001, upper bound 0.001, annealing to $10^{-5}$ in the last 10\% of training steps (see Algorithm~\ref{onecycle})
    \item \textbf{Warm restarts}: lower bound 0.0001, upper bound 0.001, cycle length = one epoch (see Algorithm~\ref{warm})
\end{itemize}

\begin{figure}[h]
\begin{algorithm}[H] 
\captionsetup{font=small}
\caption{One Cycle Policy for CapsNet}
\label{onecycle}
\begin{algorithmic}[1]
\begin{algsize}
\LineComment{OCP stands for One-Cycle Policy}
\Procedure{OCP}{$lr_{min},\ lr_{max},\ TotalSteps,\ Tcurr$}
\State $t_m \gets 0.45 \cdot TotalSteps$
\State $m \gets \frac{lr_{max}-lr_{min}}{t_m}$
\State $m_{ann} \gets 9 \cdot \frac{lr_{min}}{TotalSteps}$
\If{$Tcurr \leq t_m$}
\State $lr \gets mx + lr_{min}$
\ElsIf {$t_m \leq Tcurr \leq 2t_m$}
\State $lr \gets -m \cdot (x-2t_m) + lr_{min}$
\Else 
\State $lr \gets -m_{ann} \cdot (x-2t_m) + lr_{min}$
\EndIf
\EndProcedure
\vspace*{0mm}
\end{algsize}
\end{algorithmic}
\end{algorithm}
\vspace*{-10pt}
\end{figure}

\begin{figure}[h]
\begin{algorithm}[H] 
\captionsetup{font=small}
\caption{Warm Restarts for CapsNet: the learning rate is decayed with cosine annealing.}
\label{warm}
\begin{algorithmic}[1]
\begin{algsize}
\LineComment{WR stands for WarmRestarts}
\Procedure{WR}{$lr_{min}, lr_{max}, T_{curr}, T_i$}
\LineComment{Learning rate update}
\State $lr \gets lr_{min} + \frac{1}{2} \left(lr_{max} - lr_{min}\right) \left( 1 + \cos{\pi \frac{T_{curr}}{T_i}} \right)$
\If{$T_{curr} = T_i$} 
\LineComment{Warm Restart after $T_i$ training steps}
\State $T_{curr} \gets 0$
\Else
\LineComment{Current step update}
\State $T_{curr} \gets T_{curr} + 1$
\EndIf
\State \textbf{return} $T_{curr}$
\EndProcedure
\vspace*{0mm}
\end{algsize}
\end{algorithmic}
\end{algorithm}
\end{figure}

\subsection{Batch Size}
To realize \textit{adaptive batch size}, the batch size is set to 1 for the first 3 epochs, and then increased for 3 times every 5 epochs. That is, the user can choose a value $P$ and the batch size will assume the values $2^P$, $2^{P+1}$ and $2^{P+2}$ (see Algorithm~\ref{adabatch}). 

\begin{figure}[h]
\begin{algorithm}[H] 
\captionsetup{font=small}
\caption{AdaBatch for CapsNet: the Batch Size is increased during training.}
\label{adabatch}
\begin{algorithmic}[1]
\begin{algsize}
\Procedure{AdaBatch}{$P, CurrentEpoch$}
\If{$CurrentEpoch \leq 3$}
\State $BatchSize \gets 1$
\ElsIf{$4 \leq CurrentEpoch \leq 8$}
\State{$BatchSize \gets 2^P$}
\ElsIf{$9 \leq CurrentEpoch \leq 13$}
\State{$BatchSize \gets 2^{P+1}$}
\Else
\State{$BatchSize \gets 2^{P+2}$}
\EndIf
\EndProcedure
\vspace*{0mm}
\end{algsize}
\end{algorithmic}
\end{algorithm}
\end{figure}

\subsection{Complexity of the CapsNet Decoder}
\label{sec:decoder}
\begin{figure}[h]
\vspace*{3.5mm}
\centering
\includegraphics[width=\linewidth]{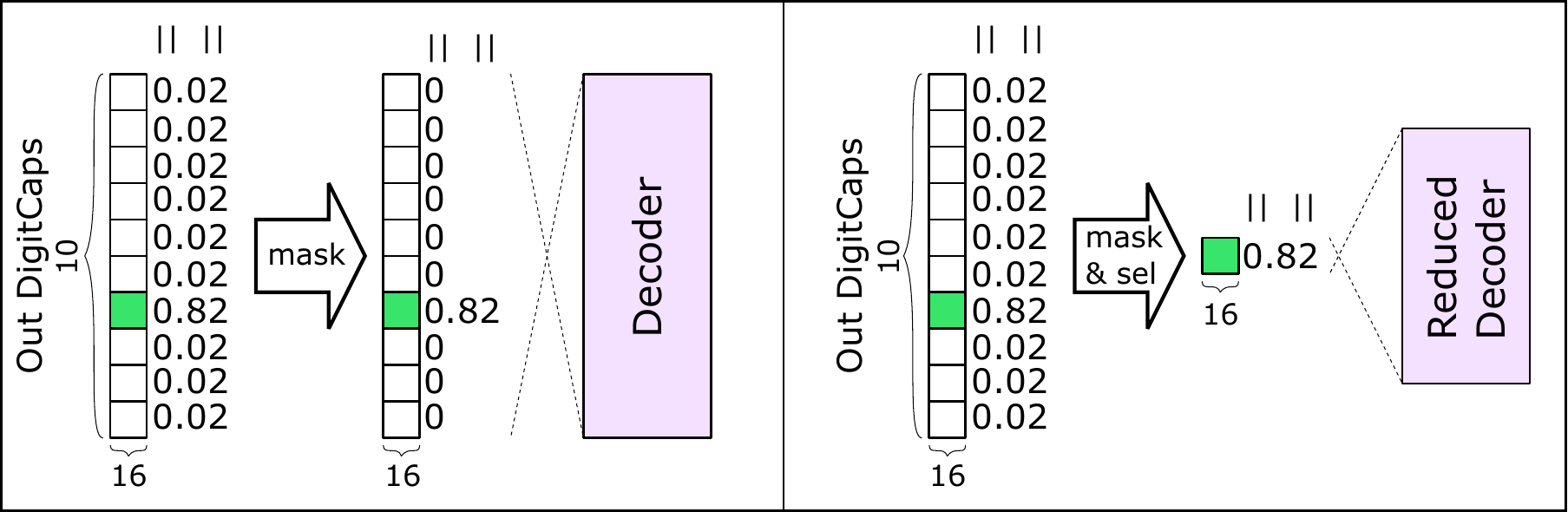}
\caption{\textbf{(left)} All the DigitCaps outputs except the one with the highest magnitude are set to zero. Then the decoder receives 10x16 inputs.\\
\textbf{(right)} Only the DigitCaps output with the highest magnitude is fed to a reduced decoder, with 1x16 inputs.}
\label{fig:decoder}
\vspace*{-10pt}
\end{figure}

The decoder is an essential component of the CapsNet. Indeed, the absence of a decoder would result in a lower accuracy of the CapsNet. The outputs of the DigitCaps layer are fed to the decoder: the highest valued vector (capsule) at the output is left untouched, while the remaining 9 vectors are set to zero (Fig.~\ref{fig:decoder} left). Thus, the decoder receives 10$\times$16 values, where 9$\times$16 are null. Therefore, we optimize the model by using a reduced-sized decoder (Fig.~\ref{fig:decoder} right) with only the 1$\times$16 inputs, which are linked to the capsule that outputs the highest probability. Overall, the original decoder has 1.4M parameters (weights and biases), while the reduced decoder provides a 5\% reduction, with 1.3M parameters. 

\subsection{Complexity Reduction through Weight Sharing}
\label{sec:weight_sharing}
The Algorithm~\ref{weightsharingpseudo} illustrates how to share the weights between the PrimaryCaps and the DigitCaps layers, by having a single tensor weight associated to all the 8-element vectors inside each 6x6 capsule. Using this method, it is possible to reduce the total number of parameters by more than 15\%, from 8.2 millions to 6.7 millions. However, the accuracy drops by almost 0.3\%, when comparing it to the baseline CapsNet.

\begin{figure}[h]
\begin{algorithm}[H]
\captionsetup{font=small}
\caption{Weight Sharing for CapsNet, applied only to the DigitCaps layer.}
\label{weightsharingpseudo}
\begin{algorithmic}[1]
\begin{algsize}
\LineComment{BatchSize is the dimension containing single elements}
\LineComment{Input size is [BatchSize, 32, 36, 8]}
\Procedure{DigitCaps}{$input, BatchSize$}
\LineComment{Weight size is [BatchSize, 32, 1, 10, 16, 8]}
\State $initialize ~ weight$
\LineComment{Bias size is [BatchSize, 1, 10, 16, 1]}
\State $initialize ~ bias$
\LineComment{We move along the dimension with 36 elements}
\LineComment{S here stands for Step}
\For{$S=1, 36$}
\LineComment{Result size is [bs, 32, 36, 10, 16, 1]}
\LineComment{We use the same weight, instead of cycling}
\State $u[S] \gets matrix\_multiply(weight[1], input[S])$
\EndFor
\LineComment{Output size is [BatchSize, 1, 10, 16, 1]}
\State $v \gets routing(u, bias)$
\State \textbf{return} $v$
\EndProcedure
\vspace*{0mm}
\end{algsize}
\end{algorithmic}
\end{algorithm}
\end{figure} 

\subsection{WarmAdaBatch}
\label{sec:output_generation}
Among the explored learning rate policies, the \textit{warm restarts} guarantees the most promising results in terms of accuracy, while the \textit{adaptive batch size} provides a good trade-off to obtain fast convergence. We propose \textit{WarmAdaBatch} (see Algorithm~\ref{warmada}), a hybrid learning rate policy to expand the space of the solutions by combining the best of the two worlds. For the first three epochs, the batch size is set to 1, then it is increased to 16 for the remaining training time. A first cycle of \textit{warm restarts} policy is done during the first three epochs, and a second one during the remaining training epochs. The learning rate variation of the \textit{WarmAdaBatch} is shown in Fig.~\ref{fig:learning_rate_warmadabatch}.

\begin{figure}[h]
\vspace*{-5pt}
\begin{algorithm}[H] 
\captionsetup{font=small}
\caption{Our WarmAdaBatch for CapsNet.}
\label{warmada}
\begin{algorithmic}[1]
\begin{algsize}
\LineComment{WAB stands for WarmAdaBatch}
\Procedure{WAB}{$lr_{min}$, $lr_{max}$, $MaxEpoch$, $MaxStep$}
\State $T_{curr} \gets 0$
\For{$Epoch=1, MaxEpoch$}
\LineComment{Batch size update}
\State {$Adabatch(4,Epoch)$}
\If{$Epoch \leq 3$}
\LineComment{Steps in 3 epochs with batch size 1}
\State{$T_i \gets 3 * 60,000$}
\Else
\LineComment{Steps in 27 epochs with batch size 16}
\State{$T_i \gets 27 * 3,750$}
\EndIf 
\For{$Step=1, MaxStep$}
\LineComment{Learning Rate update}
\State $T_{curr} \gets WR(lr_{min}, lr_{max}, T_{curr}, T_{i})$
\EndFor
\EndFor
\EndProcedure
\end{algsize}
\end{algorithmic}
\end{algorithm}
\end{figure}

\subsection{Optimization Choices}
Our framework is able to automatically optimize CapsNets and its training depending on the parameters that a user wants to improve. For instance,  using \textit{WarmAdaBatch}, the accuracy and the training time are automatically co-optimized. The number of parameters can be reduced, at the cost of some accuracy loss and training time increase, by enabling the \textit{weight sharing}, along with the \textit{WarmAdaBatch}.

\section{Evaluation}
\label{sec:evaluation}

\subsection{Experimental Setup}
\begin{figure*}[t]
\vspace*{0mm}
    \centering
    \begin{minipage}[c]{\linewidth}
    \begin{figure}[H]
    \centering
    \includegraphics[width=.82\textwidth]{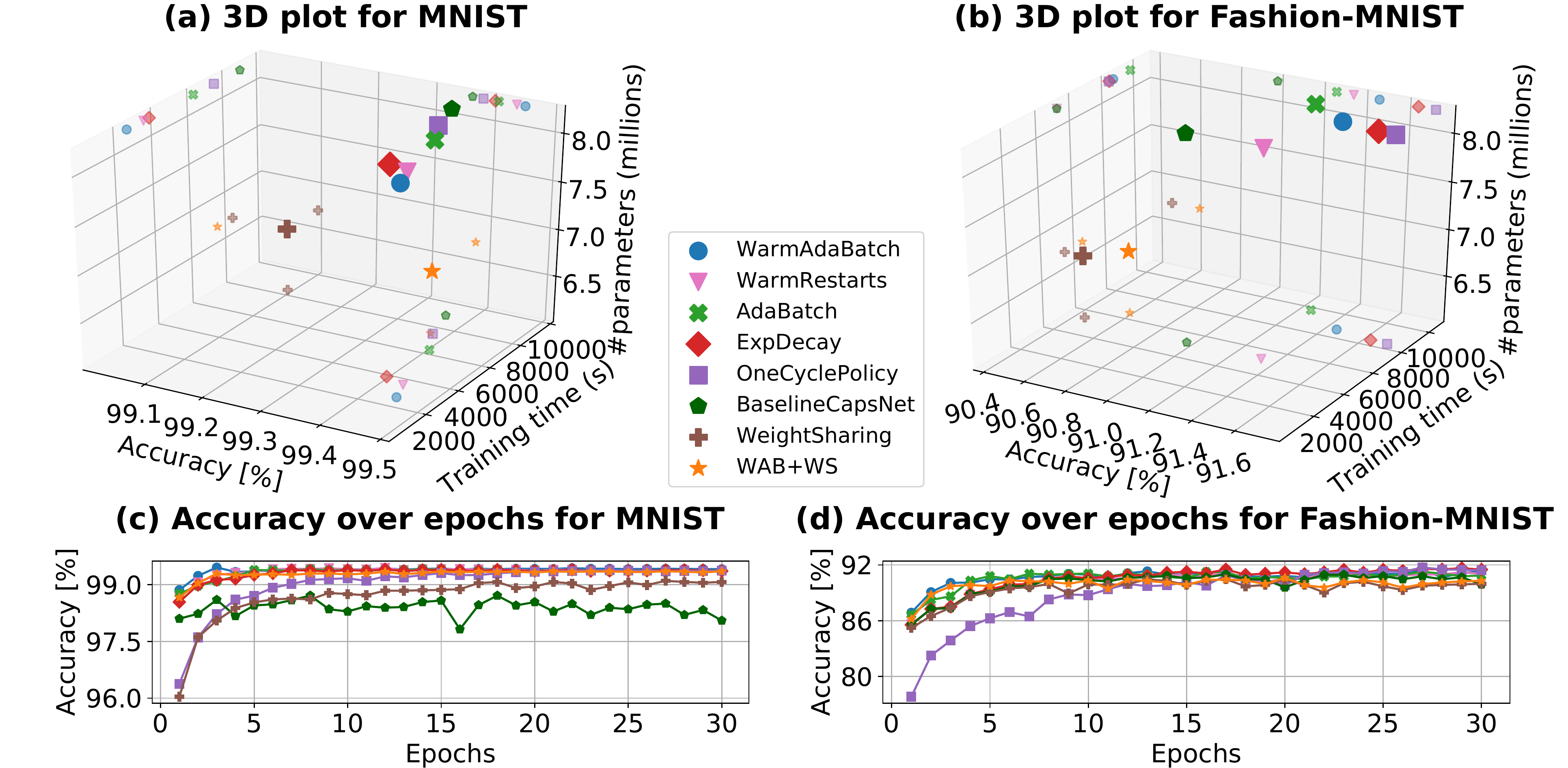}
    \caption{\textit{The legend is common for all the figures.} \textbf{(a,b)} Comparison of different optimization types integrated in our \textit{FasTrCaps} framework, on the basis of accuracy, training time and number of parameters. The training time is computed as the number of epochs to reach the maximum accuracy, multiplied by time (in seconds) per epoch. The abbreviated terms WAB and WS stand for \textit{WarmAdaBatch} and \textit{WeightSharing}, respectively, with \textit{WeightSharing} including also the small-decoder optimization. \textbf{(c,d)} Accuracy improvements / changes over the training epochs for different optimization solutions. 
    (\textbf{a,c}) Results for MNIST. (\textbf{b,d}) Results for Fashion-MNIST.}
    \label{fig:evaluation_comparison}
    \end{figure}
    \end{minipage}
\end{figure*}

We developed our framework using the PyTorch library~\cite{paszke2017automatic}, running on two Nvidia GTX 1080 Ti GPUs. We tested it on the MNIST~\cite{mnist} and Fashion-MNIST~\cite{Fashion-MNIST} datasets. Both datasets are composed of 60,000 samples for training and 10,000 test samples each. The MNIST is a collection of handwritten digits, while the Fashion-MNIST is a collection of grayscale fashion products. After each training epoch, a test is performed. At the beginning of each epoch, the samples for training are randomly shuffled, while the testing samples are kept in the same order. The accuracy values are computed by averaging 5 training runs. Each training run lasts for 30 epochs, with the settings for each policy equal to the ones described in Section~\ref{sec:methodology}. The results are shown in Table~\ref{tab:FashionMNIST} and Fig.~\ref{fig:evaluation_comparison}a,c.


\begin{table}[h]
\vspace{2mm}
    \captionsetup{font=small}
    \centering
    \caption{Accuracy results obtained with CapsNet for the Fashion-MNIST dataset, applying different proposed solutions.}
    \label{tab:FashionMNIST}
    \resizebox{\linewidth}{!}{%
    \begin{tabular}{|c|c|c|c|c|c|}
    \hline
    \multicolumn{2}{|c|}{\textbf{Accuracy}} & \multicolumn{2}{c|}{\textbf{\begin{tabular}[c]{@{}c@{}}Epochs to reach\\ max accuracy\end{tabular}}} & \multirow{2}{*}{\textbf{Parameters}} & \multirow{2}{*}{\textbf{\begin{tabular}[c]{@{}c@{}}Weight\\ Sharing\end{tabular}}} \\ \cline{1-4}
    \textit{FashionMNIST}  & \textit{MNIST} & \textit{FashionMNIST}                                & \textit{MNIST}                                &                                      &                                                                                    \\ \hline
    90.99\%                & 99.37\%        & 17                                                   & 29                                            & Fixed (Baseline)                             & No                                                                                 \\ \hline
    91.47\%                & 99.45\%        & 27                                                   & 8                                             & WAB                                  & No                                                                                 \\ \hline
    90.47\%                & 99.26\%        & 17                                                   & 26                                            & Fixed (Baseline)                            & Yes                                                                                \\ \hline
    90.67\%                & 99.38\%        & 20                                                   & 11                                            & WAB                                  & Yes                                                                                \\ \hline
    \end{tabular}
    }
\end{table}

\subsection{Accuracy Results for the MNIST dataset}
\textbf{Evaluating the Learning Rate Policies}: Among the state-of-the art learning policies that we enhanced for CapsNets, the \textit{warm restarts} is the most promising one, as the maximum accuracy improved by 0.074\%. CapsNet with \textit{warm restarts} reaches the maximum accuracy of the baseline (with fixed learning rate) in 6 epochs rather than in 29 epochs as required by the baseline, thereby providing a training time reduction of 62.07\%. 

\textbf{Evaluating Adaptive Batch Size}: Different combinations of batch sizes in \textit{adaptive batch size} algorithm have been tested, since the smaller the batch size is, the faster the initial convergence. However, large batch sizes lead to slightly higher accuracy after 30 epochs and, most importantly, to a reduced training time. In fact, a CapsNet training epoch with batch size 1 lasts for 7 minutes, while with batch size 128 it lasts for only 28 seconds. Batch size 16 is a good trade off between fast convergence and short training time (i.e., 49 sec/epoch). 
The best results, applying \textit{adaptive batch size}, are obtained using batch size 1 for the first three epochs, and then increasing it to 16 for the remaining part of the training. With this parameter selection, there is a 0.04\% accuracy gain w.r.t. the baseline, and the maximum accuracy of the baseline is reached in 5 epochs rather than in 29 epochs as required by the baseline. However, the first three epochs take a longer time (88\% longer time) because of the reduced batch size, so \textit{adaptive batch size} alone is not convenient. However, the total training time is reduced by 30\% as compared to the baseline.

\textbf{Evaluating WarmAdaBatch}: As for the batch, the first cycle of learning rate lasts for 3 epochs and the second one for 27 epochs. Variation of batch size and learning rate cycles are synchronized. This solution allows to have a 0.088\% gain in accuracy w.r.t. the baseline CapsNet implementation, and the baseline maximum accuracy is reached by CapsNet with \textit{WarmAdaBatch} in 3 epochs against 29 epochs. After the first three epochs, the batch size changes and the learning rate is restarted. Hence, there is a drop of accuracy, which re-converges in a few steps to the highest and stable value. 

\textbf{Evaluating Weight Sharing}: By applying \textit{weight sharing} to the DigitCaps layer, we are able to achieve a 15\% reduction in the number of total parameters, decreasing from 8.2 millions to 6.7 millions. However, this reductions also leads to a slight decrease in the maximum accuracy, i.e., by 0.26\%.

\subsection{Comparison of Different Optimization Types}

On the CapsNet model with the MNIST dataset, we also compare the different types of optimizations in terms of accuracy, and based on the training time to reach the maximum accuracy and the number of parameters. As we can see in Fig.~\ref{fig:evaluation_comparison}a, we compare different optimization methods in a 3-dimensional space. This representation provides the Pareto-optimal solutions, depending on the optimization goals. We also compare, in Fig.~\ref{fig:evaluation_comparison}c, the accuracy and the learning rate evolution in different epochs, for \textit{AdaBatch}, \textit{WarmRestarts} and \textit{WarmAdaBatch}. Among the space of the potential solutions, we discuss the following two Pareto-optimal choices in detail, i.e., the \textit{WarmAdaBatch} and the combination of \textit{WarmAdaBatch} and \textit{weight sharing}, which we call WAB+WS.

\textbf{WarmAdaBatch}: This solution provides the optimal point in terms of accuracy and training time, because it achieves the highest accuracy (99.45\%) in the shortest time (3 epochs). Varying the batch size boosts the accuracy in the first epoch and the restart policy contributes towards accelerating the training.

\textbf{WAB+WS}: The standalone \textit{weight sharing} reduces the number of parameters by 15\%. By a combination of it with \textit{WarmAdaBatch}, the accuracy loss is compensated (99.38\% vs. 99.37\% of the baseline), while the training time is shorter than the baseline (18 epochs vs. 29 epochs) but longer than the simple \textit{WarmAdaBatch}. Our framework chooses this solution if the reduction in the number of parameters is also included in the optimization goal.

\subsection{Accuracy Results for Fashion-MNIST}

The results for the Fashion-MNIST dataset are shown in Table~\ref{tab:FashionMNIST} and Fig.~\ref{fig:evaluation_comparison}b,d. However, while the WAB+WS policy is the most effective policy for reducing the network parameters while keeping a relatively high accuracy, not only the WAB policy but also the ExpDecay policy and the One-Cycle-Policy show good accuracy and training time results. The WAB policy is able to keep the same training time as for the best policies but at the cost of a slight accuracy loss. Hence, even though Fashion-MNIST and MNIST require an equivalent CapsNet architecture (i.e., without any changes), our WAB policy for Fashion-MNIST is comparable to other learning policies. 

\section{Conclusion}
In this paper, we proposed \textit{FasTrCaps}, a novel framework for accelerating the CapsNet training. It integrates multiple lightweight optimizations into the training loop, to reduce the training time and/or the number of parameters, based on the requirements needed. We enhanced the different learning policies, for the first time, for accelerating the training of CapsNets. Afterwards, we discussed how an integrated training framework can be developed to find the Pareto-optimal solutions, including different new optimizations for fast training like \textit{WarmAdaBatch}, complexity reduction for the CapsNet decoder, and \textit{weight sharing} for CapsNets. These solutions not only provide significant reduction in the training time while preserving or even improving the accuracy, but also enable a new mechanism to provide trade-off between training time, network complexity, and classification accuracy. This enables new design points under different user-provided constraints. The source-code of our \textit{FasTrCaps} framework can be found at \url{https://github.com/Alexei95/FasTrCaps}.

\section*{Acknowledgments}

This work has been partially supported by the Doctoral College Resilient Embedded Systems which is run jointly by TU Wien's Faculty of Informatics and FH-Technikum Wien.

\begin{refsize}
\bibliographystyle{abbrvnat}
\bibliography{main.bib}
\end{refsize}
\end{document}